\documentclass[twoside,leqno,twocolumn]{article}

\usepackage[letterpaper]{geometry}
\usepackage{ltexpprt}
\usepackage{amsmath,amssymb}
\usepackage{mathtools}
\usepackage{bbm}
\usepackage{xcolor}
\usepackage{graphicx}
\usepackage{algorithmic}
\usepackage{subcaption}
\usepackage{hyperref}

\DeclareMathAlphabet\mathbfcal{OMS}{cmsy}{b}{n}
\title{\Large Impact of Community Structure on Consensus Machine Learning\thanks{
BH and DT were supported in part by the Simons Foundation (Award \#578333) and the NSF (DMS-1551069, EDT).}}

\author{Bao Huynh\thanks{Department of Mathematics, University at Buffalo, New York, NY (baohuynh@buffalo.edu).}
\and Haimonti Dutta\thanks{Department of Mathematics, University at Buffalo, New York, NY.}\thanks{Management Science and Systems Department, University at Buffalo, New York, NY. (haimonti@buffalo.edu)}
\and Dane Taylor\thanks{Department of Mathematics, University at Buffalo, New York, NY. ({danet@buffalo.edu}).}}

\begin{document}
\maketitle

%%%%%%%%%%%%%%%%%%%%%%%%%%%%%%
%   abstract
%%%%%%%%%%%%%%%%%%%%%%%%%%%%%%
\begin{abstract}
Consensus dynamics support decentralized machine learning for  data that is distributed across a cloud compute cluster or across  the internet of things. In these and other settings, one seeks to minimize the time $\tau_\epsilon$ required to obtain consensus within some $\epsilon>0$ margin of error.  $\tau_\epsilon$ typically depends on the topology of the underlying communication network,  and for many algorithms $\tau_\epsilon$ depends on the second-smallest eigenvalue $\lambda_2\in[0,1]$ of the network's normalized Laplacian matrix: $\tau_\epsilon\sim\mathcal{O}(\lambda_2^{-1})$. Here, we analyze the effect on   $\tau_\epsilon$ of network community structure, which can arise when compute nodes/sensors are  spatially clustered, for example. We study  consensus machine learning over networks drawn from  stochastic block models, which yield random networks that can contain heterogeneous communities with different sizes and densities.  Using   random matrix theory, we  analyze the effects of communities on $\lambda_2$ and consensus, finding that $\lambda_2$ generally increases (i.e., $\tau_\epsilon$ decreases) as one decreases the extent of community structure. We further observe that there exists a critical level of community structure at which $\tau_\epsilon$ reaches a lower bound and is no longer limited by the presence of communities. We support our findings with  empirical experiments for decentralized support vector machines. 
\end{abstract}

~

\noindent{\bf Keywords}: community structure; consensus learning; support vector machines;

%%%%%%%%%%%%%%%%%%%%%%%%%%%%%%
% sections
%%%%%%%%%%%%%%%%%%%%%%%%%%%%%%

\section{Introduction}\label{sec:intro}
Wireless sensor networks (WSNs) are used extensively in many applications including habitat monitoring, smart-building monitoring, industrial automation, and target tracking \cite{Szew_04,Batra_13a,Batra_2014,Souza_16}. These ad-hoc networks present interesting problems for system design: 
on the one hand, their low cost enables distributed, massive scale   compute infrastructure; 
on the other, their limited power and low reliability suggests that system performance is enhanced if local communication is preferred  (i.e., between nearby sensors).
% communication is preferentially.
%with small number of neighbors is utilized. 
The massive scale makes it unrealistic to rely on careful placement or uniform arrangement of sensors. Instead, most applications nowadays rely on self organization and localization \cite{Collier_04,Kuria_14} of these tiny mobile devices, rather than relying on globally accessible beacons or expensive GPS to localize sensors. The structure and organizational properties of sensor networks often provide useful clues for its efficient management. The presence or absence of community structure is one such important characteristic that has received sufficient attention in literature \cite{Newman_06a,Clauset_04a}. 

In addition, there is growing interest to pair such sensors with decentralized computing nodes that can cooperatively learn and solve distributed optimization problems for machine learning
%to solve %lends them amenable to solving large scale 
%decentralized machine learning problems 
\cite{Nedic_2017,Duchi_12a,Dutta_20a}. For instance, in consensus learning, nodes can individually sense data of different modalities (such as image, audio, text) to learn local models; they can then cooperatively engage in learning a global model instead of transferring data to a beacon or a base station. The study of network-topology effects on communication-computation trade-offs for distributed optimization algorithms is an area of active research \cite{Savic_14a,Dimakis_10a,XiaoBK07,XiaoBL05,ZhuSJS20,TutunovBJ19}. The effect of network scaling for average consensus has been studied by Nedic et al.\cite{Nedic_2017} for some predefined topologies (such as two dimensional grids, star, Erdos Renyi and geometric random graphs); Duchi et al. \cite{Duchi_12a} study dual stochastic sub-gradient averaging and show that the convergence rate scales inversely as a function of the spectral gap of the network. However, it remains much less studied how community structure within random networks impacts the convergence properties of decentralized algorithms for distributed machine learning and optimization.

Focusing on consensus-based machine-learning algorithms, we study the impact of community structure on   \emph{convergence time} $\tau_\epsilon$:  the time required for the states of all nodes to converge  within some small threshold $\epsilon>0$ of their limit.
Community structure can inherently arise when compute nodes are physically clustered together, e.g., by being located within the same building, room, etc.
It is well known that   $\tau_\epsilon\sim\mathcal{O}(\lambda_2^{-1})$ for many consensus-based algorithms, where $\lambda_2\in(0,1)$ is the second-smallest eigenvalue  of a normalized Laplacian matrix.
At the same time,   $\lambda_2$ has been extensively studied in the context of graph cuts and spectral clustering  \cite{chung1996laplacians}.  For example, one can bound $\lambda_2\le2h$  using the Cheeger constant $h$, which equals zero 
%$h = \min_{X\subset \mathcal{V}} \partial(X)/\min(\text{vol}(X),\text{vol}(\mathcal{V}\setminus X))$, where $\partial(X)$  and $\text{vol}(X)$ refer to the boundary and volume of the subgraph restricted to nodes $X\subset \mathcal{V}$.  Since $h=0$ 
if a community is disconnected from the remaining network
%, $\lambda_2\to0$ is an expected limit under very strong community structure 
(in which case $\tau_\epsilon$ diverges). Understanding the effects of community structure on $\lambda_2$ in a broader setting (such as random networks containing communities of different sizes and densities), as well as its effect on consensus machine learning, remains underexplored.

In this paper, we take an important step in addressing this gap in literature. 
We study  consensus machine learning over stochastic block models, a popular random-network model that allows for heterogeneous communities with diverse sizes and densities.
Using recently developed techniques to characterize the spectral decompositions of low-rank-perturbed random matrices, we analyze the effect of communities on $\lambda_2$ and consensus.
Our main finding is that decreasing the extent/prevalence of community structure lowers convergence times, however there exists a lower bound on convergence time that does not depend on community structure. The practical consequence is that one often can decrease $\tau_\epsilon$ by decreasing the   extent of community structure, but this is effective only to a point.
%but there is a lower bound for $\tau_\epsilon$ that is   independent of the  community structure.
%
We support our theoretical findings with  numerical experiments for   ``vanilla'' consensus dynamics and decentralized learning with  consensus-based  support vector machines.

This paper  is organized  as follows: 
(Sec.~\ref{sec:back}) presents background information; 
%on consensus, community structure, and random-matrix theory;
(Sec.~\ref{sec:consensus}) discusses  main  findings for how community structure impacts consensus; 
(Sec.~\ref{sec:gadget}) presents empirical findings for decentralized support vector machines; and
(Sec.~\ref{sec:conclusion}) has concluding remarks.

\section{Background Information}\label{sec:back}

We will develop random matrix theory to characterize $\lambda_2$ and $\tau_\epsilon$ for consensus machine learning over a popular model for random networks with community structure. Here, we provide the necessary background information:  
(Sec.~\ref{sec:vanilla}) consensus dynamics over a network;
(Sec.~\ref{sec:SBM}) the stochastic block model for random networks with heterogeneous communities; 
and (Sec.~\ref{sec:RMT}) spectral  theory for large block random matrices.

\subsection{``Vanilla'' Consensus over Networks.}\label{sec:vanilla}

We   first study consensus dynamics in its most elementary form: the consensus of scalar data over a network. Despite its simplicity, it is also  a common framework when using a sensor network to reliably extract, for example, the speed of a moving vehicle \cite{Saeednia_17a} or the temperature of a room, fluid or object \cite{Kim_15a}.  Later in Sec.~\ref{sec:gadget}, we will study the consensus of support vector machines, and in that case, one seeks to obtain a consensus of locally learned  models for  distributed data (with the consensus and learning dynamics occurring together  in tandem).

Consider a communication network consisting of a set  $\mathcal{V}=\{1,\hdots, n \}$ of nodes (each   representing a computing element) and  a set of undirected, unweighted edges $\mathcal{E} = \{ (i,j)\} \in \mathcal{V} \times \mathcal{V} $ (each representing a pairwise communication between compute nodes).  The topology of network communication is formally represented by a  graph  $G(\mathcal{V},\mathcal{E})$, which can be  equivalently encoded by an adjacency matrix $\textbf{A}\in\mathbb{R}^{n \times n}$ with entries $A_{ij} = 1$ if $(i,j)\in\mathcal{E}$ and $A_{ij} =0$ otherwise. Letting $d_i=\sum_j A_{ij}$ be the degree of each node $i$ and $\textbf{D} =\text{diag}[d_1,\dots,d_N]$, we can define several other   graph-encoding matrices that are associated with $G(\mathcal{V},\mathcal{E})$:  a transition matrix $\textbf{P} =  \textbf{D}^{-1} \textbf{A}$ (e.g., for a discrete-time Markov chain);
an unnormalized Laplacian matrix $ \textbf{L} = \textbf{D}  -   \textbf{A} $; and a normalized Laplacian matrix  $\widehat{\textbf{L}} = \textbf{I} - \textbf{D}^{-1/2} \textbf{A} \textbf{D}^{-1/2}$.  (Matrix $\textbf{I}$ is the size-$n$ identity matrix.) 

By construction, $\textbf{P} $ is a row-stochastic matrix, implying that   $\textbf{1} = [ 1\dots,1]^T$ is a right eigenvector with eigenvalue $\lambda_1=1$.  We denote the associated left eigenvector $\pi = [\pi_{1} ,\dots,  \pi_{n}]^T$, which is normalized in 1-norm so that $\sum_i \pi_i = 1$. Because $\textbf{P} $ is nonnegative, the Perron-Frobenius theorem for nonnegative matrices ensures that $\lambda_1\ge |\lambda_i|$ for any $i\in\mathcal{V}$, and that the entries satisfy $\pi_{i}\ge 0$. If the graph is strongly connected, then $\textbf{P} $ is an irreducible matrix, and one additionally has $\pi_{i}> 0$.

Another consequence of   $\textbf{P} $ being a row-stochastic matrix is that for any vector ${\bf x}\in\mathbb{R}^n$,  the operation $\textbf{P} {\bf x}$ implements for each node $i$ an averaging of the $x_j$ values across its set of neighboring nodes:  $\mathcal{N}(i)=\{j|A_{ij}>0\}$.
That is,   
$[\textbf{P} {\bf x}]_i = \sum_j P_{ij} x_j =   \frac{1}{|\mathcal{N}(i)|} \sum_{j\in\mathcal{N}(i)} x_{j}$, which uses $d_i=|\mathcal{N}(i)|$. This local-averaging process can be iteratively applied to yield the  synchronous consensus algorithm for scalars
\begin{align} \label{eq:scalar}
	\textbf{x}(t+1) = \textbf{P} \textbf{x}(t).
\end{align} 
Here, $\textbf{x}(t) = [ x_{1}(t) ,\dots, x_{n}(t)]^T \in \mathbb{R}^{n}$ and each   $x_{i}(t) \in \mathbb{R}$ gives the state of node $i$ at iteration $t=0,1,2,\dots$.
Because ${\bf 1}$ is a right eigenvector, $\textbf{P}{\bf 1} ={\bf 1}$, any constant-valued function ${\bf x}^* = \overline{x} {\bf 1}$ for some scalar $\overline{x}\in\mathbb{R}$ is a fixed-point solution for the discrete-time dynamics given by Eq.~\eqref{eq:scalar}. 
When $\textbf{P}$ is a \emph{primitive matrix}, i.e., there exists a $t^*>0$ such that $[\textbf{P}^t]_{ij}>0$ for all $t>t^*$, then ${\bf x}^*=\overline{x} {\bf 1}$ is a globally attracting fixed-point solution with $\overline{x} = \langle {\bf x}(0), {\bf \pi}\rangle $. Primitive matrices are equivalent to irreducible aperiodic nonnegative matrices, which can be proven by making various topological assumptions about the graph such as it being strongly connected (which ensures irreducibility) and containing at least one self edge (which ensures aperiodicity).

The convergence  ${\bf x}(t) \to{\bf x}^*$ can be characterized in several ways including the \emph{convergence rate}
\begin{align}
    r \equiv \sup_{{\bf x}(0)\in\mathcal{S}} \lim_{t\to\infty} \left(\frac{||{\bf x}(t) - {\bf x}^*||}{||{\bf x}(0) - {\bf x}^*||} \right)^{1/t},
\end{align}
where $\mathcal{S}=\text{span}({\bf 1})$,
and \emph{convergence time}
\begin{align}
    \tau_\epsilon \equiv \min \left\{ t^* \text{ s.t. }\frac{||{\bf x}(t) - {\bf x}^*||_\infty}{||{\bf x}(0) - {\bf x}^*||_\infty} \le \epsilon ~\forall t\ge t^* \right\}.
\end{align}
The convergence rate can be shown to satisfy $r =  |\mu_2|$, where we have ordered the eigenvalues of $\textbf{P}$ as  $\mu_1>|\mu_2|\ge|\mu_{3}|\ge\dots$. If we assume $\textbf{P}$ is diagonalizable, $\textbf{P} =  \sum_{i=1}^n \mu_i {\bf w}_i{\bf v}_i^T$ with associated left and right eigenvectors ${\bf w}_i$ and ${\bf v}_i$, with ${\bf w}_1=\pi$ and ${\bf v}_1 ={\bf 1}$, then we can expand  
$\textbf{x}(0) = \sum_{j=1}^{n} a_{j} \textbf{v}^{(j)}$ with $a_j = \langle {\bf w}_j ,{\bf x}(0)\rangle$ to obtain the solution $\textbf{x}(t) = \sum_{j=1}^{n} \mu_j^t a_{j} \textbf{v}^{(j)}$. Using $\mu_1 = 1$, and after some rearranging, one obtains the bound
\begin{align} 
	||   \textbf{x}(t) - \overline{x} \textbf{1}  || &\leq \sum_{j=2}^{n} |a_{j}| |\mu_{j}|^{t}  || \textbf{v}^{(j)} || \nonumber\\
	&\leq  |\mu_{2}|^{t}    || \textbf{x} (0) - \textbf{x}^* || .
\end{align} 
 To obtain a bound on $\tau_\epsilon$, we identify the minimum time $t$ such that $|\mu_{2}|^{t} \le \varepsilon$, for which it is straightforward to obtain $\tau_\epsilon \le \ln( \varepsilon ) /\ln( |\mu_{2}| )$. 
Using a Taylor series expansion, one can also obtain the first-order bound 
$\tau_\epsilon \le \ln( \varepsilon ) /(1- |\mu_{2}| )$.

Finally, it is worth noting that there is a one-to-one mapping between the eigenvalues of ${\bf P}$ and $\widehat{{\bf L}}$: for each eigenvalue $\lambda_j$ of $\widehat{{\bf L}}$ with  eigenvector ${\bf u}_j$, $\mu_j = 1-\lambda_j$ is an eigenvalue of  ${\bf P}$ with left and right eigenvectors given by ${\bf w}_j ={\bf D}^{1/2}{\bf u}_j$ and ${\bf v}_j ={\bf D}^{-1/2}{\bf u}_j$, respectively. Therefore, the first-order bound can be equivalently expressed using the smallest nonzero eigenvalue $\lambda_2$ of the normalized Laplacian:  $\tau_\epsilon = \mathcal{O}(1/ \lambda_2 ).$ (Note that we have additionally assumed that the  eigenvalue that has second-largest magnitude is positive. This is true for many networks, including all the experiments that we have conducted.)

\begin{figure}[t]
  \centering
    \includegraphics[width=.4\linewidth]{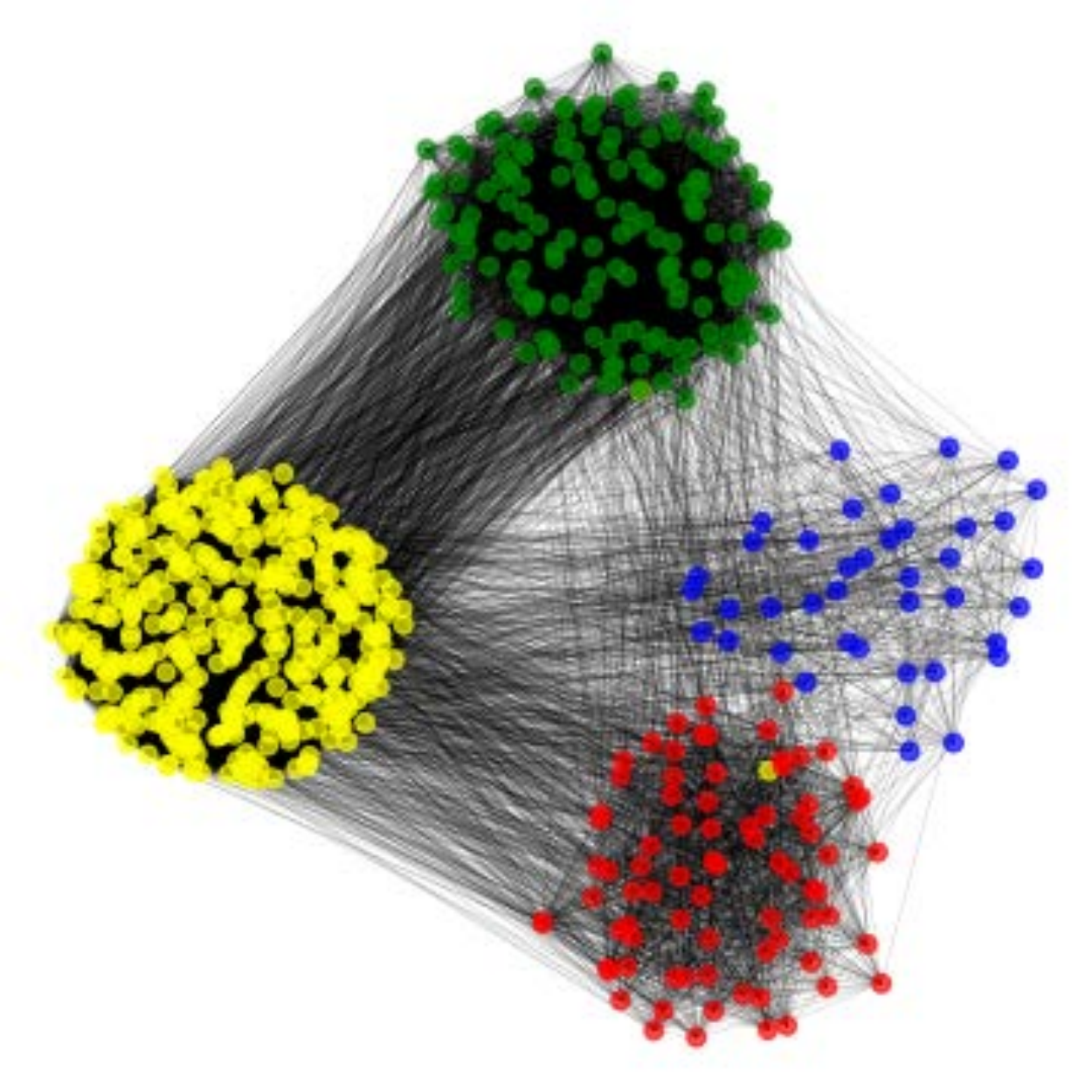}
    \includegraphics[width=.56\linewidth]{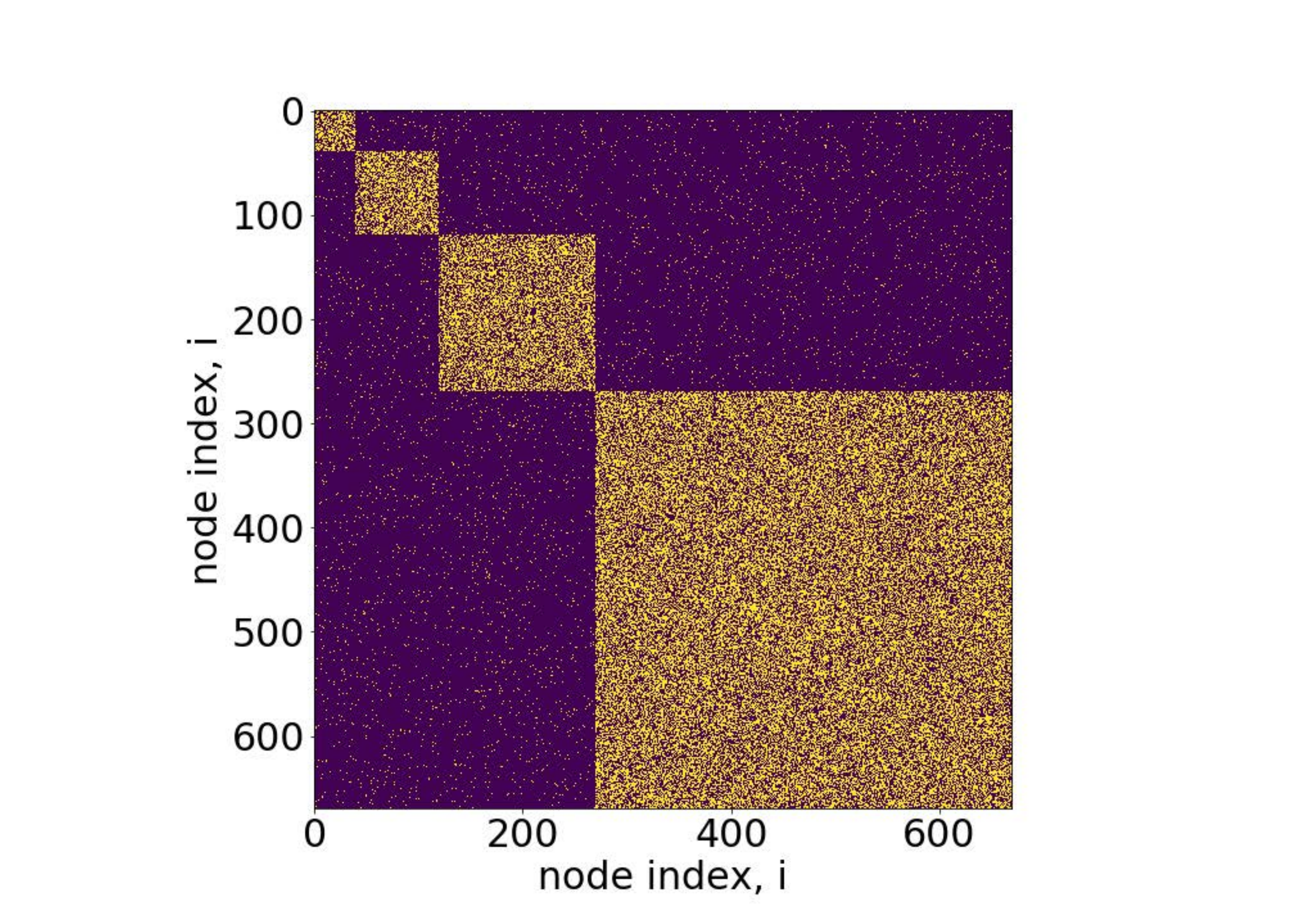}
  \caption{
  (left) Network sampled form a stochastic block model (SBM) with $K=4$ communities of sizes $[n_1,n_2,n_3,n_4]=[40,80,150,400]$ and the within/outside edge probabilities are given by $[p_{in},p_{out}]=[0.5,0.02]$.
  (right) A visualization of the corresponding block-structured adjacency matrix ${\bf A}$. Yellow dots indicate nonzero entries (i.e., edges).
  }
  \label{fig:f1}
\end{figure}

%%%%%%%%%%%%%%%%%%%%%%%%%%%%%%
%%%%%%%%%%%%%%%%%%%%%%%%%%%%%%

\subsection{Random Networks with Communities.}\label{sec:SBM}
We study consensus machine learning over stochastic block models (SBMs), which are a popular stochastic-generative model for random graphs that contain communities. A network is sampled from an SBM as follows:
the set  $\mathcal{V} = \{ 1, \dots ,n \}$ of vertices is partitioned into $K$  disjoint subsets $\{ \mathcal{V}_{1}, \dots, \mathcal{V}_{K} \}$ known as \emph{communities} or \emph{blocks}. We denote their sizes $n_r = |\mathcal{V}_r|$, and we define the matrix $\textbf{N} = \text{diag}[ n_{1}\dots, n_{K}] $. We define $n=\sum_r n_r$. For each node $i\in\mathcal{V}$, we denote by $c_i$ the community to which it belongs. Next, for each pair  $i,j\in\mathcal{V}\times \mathcal{V}$ of nodes, we create an edge $(i,j)$ as an i.i.d. boolean random variable with probability $\Pi_{c_ic_j}$. That is $\boldsymbol \Pi \in\mathbb{R}^{K\times K}$ is an edge-probability matrix in which each entry $\Pi_{rs}$ indicates the probability of an edge between a node in community $r$ and one in $s$.

\begin{figure*}[t]
  \centering 
    \includegraphics[width=.4\linewidth]{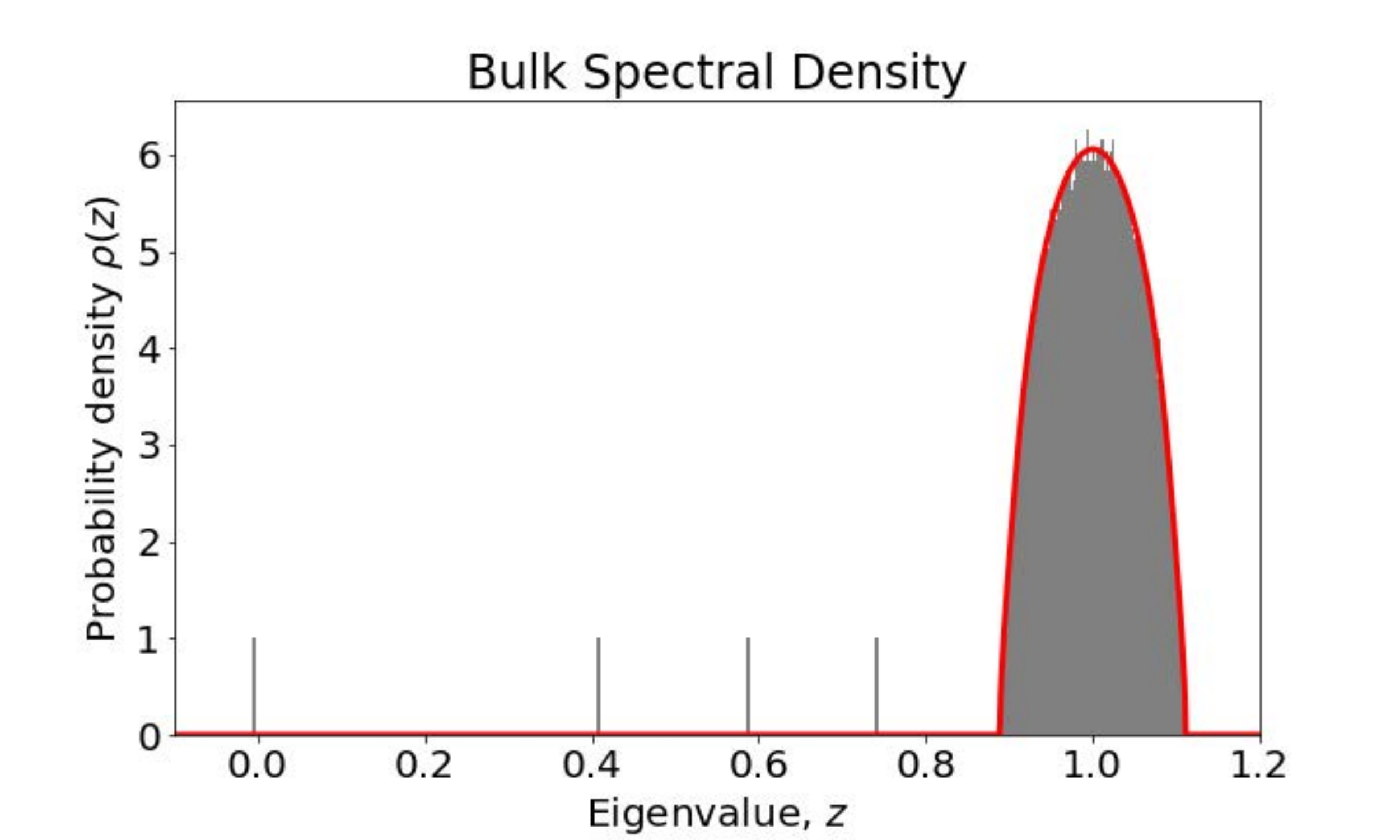}
    \includegraphics[width=.48\linewidth]{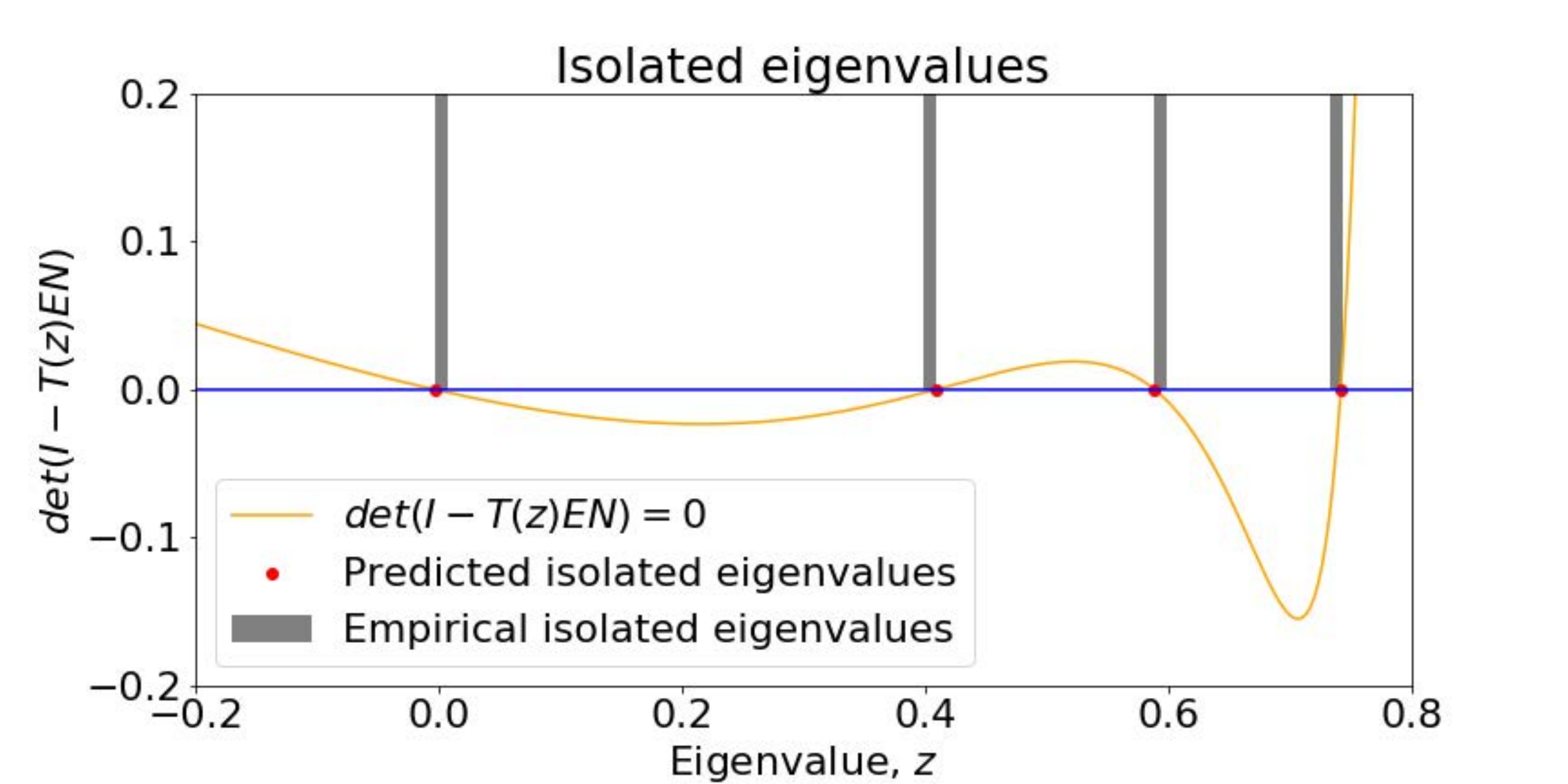}
    \caption{
    Eigenvalues of a normalized Laplacian $\widehat{{\bf L}}$ for a network that is sampled from an SBM with $n=7000$ nodes, $k=4$ communities with sizes $[n_1,n_2,n_3.n_4]=[500,1000,2000,3500]$, and $[p_{in},p_{out}]=[0.1,0.02]$.
    There are two types of eigenvalues:
    (left) \emph{Bulk eigenvalues} whose empirical density converges  a distribution $\rho(\lambda)$ as $n\to\infty$, which  can be solved using Eq.~\eqref{eq:Stieltjes} (red curve). The left and right boundaries of the support $\text{supp}(\rho)=[\lambda_L,\lambda_R]$ can be solved using Eq.~\eqref{eq:Jacobian}. The four vertical bars on the left of $\rho(\lambda)$ are the \emph{isolated eigenvalues} $\{\lambda_i\}$.
    (right) A ``zoom in'' on the isolated eigenvalues, which can be solved using Eq.~\eqref{eq:iso_2} and which have a one-to-one correspondence to the eigenvalues of the expected Laplacian $\mathbb{E}[\widehat{{\bf L}}]$.   
    }
  \label{fig:RMT}
\end{figure*}

In our experiments, we will focus on   the special case where the edge probabilities are one of two choices
\begin{align}
{ \Pi}_{rs} = \begin{cases} p_{in}, \text{if $r=s$} \\ p_{out}, \text{if $r \neq s$}. \end{cases}
\end{align}
In this case, resulting model can be fully specified by the community sizes $\{n_r\}$ along with the \emph{within-community} and \emph{between-community edge probabilities}, $p_{in}$ and $p_{out}$, respectively. We finally define the \emph{community prevalence} $\Delta = p_{in}-p_{out}$.

SBMs define a stochastic-generative model for random graphs containing heterogeneous communities with tunable properties; however, one can equivalently consider them as a model for block-structured random matrices. In particular, each entry $A_{ij}\in\{0,1\}$ is an independent Bernoulli random variable with probability $\Pi_{c_ic_j}$, and so each network that is sampled from an SBM has a one-to-one correspondence to a random   matrix in which there are blocks of matrix entries that have common statistical properties. Because of the block structure, these properties can be represented using low-rank matrices. For example,  the expectation and variance of an adjacency matrix across the ensemble are given by $\mathbb{E}[A]={\bf S}\Pi {\bf S}^T$ and 
$\mathbb{VAR}[A]={\bf S}[{\bf \Pi}\circ(1-{\bf \Pi})] {\bf S}^T$, respectively, where ${\bf S}\in\mathbb{R}^{n\times K}$ is a binary-valued matrix that provides a \emph{one-hot encoding} for which node belongs to which community: $S_{ir} = 1$ if $i\in\mathcal{V}_r$ and $S_{ir} = 0$ otherwise. (The symbol $\circ$ indicates the Haddamard product, or entrywise multiplication.) The normalized Laplacian matrix $\widehat{{\bf L}}$ can be similarly characterized, and in the following section we present random matrix theory to predict,  in expectation, the eigenvalues of $\widehat{{\bf L}}$  for the large-$n$ limit (including, but not limited to, $\lambda_2$).

%%%%%%%%%%%%%%%%%%%%%%%%%%%%%%
%%%%%%%%%%%%%%%%%%%%%%%%%%%%%%
\subsection{Random Matrix Theory for SBMs.}\label{sec:RMT}
There is a growing literature developing random matrix theory for SBMs, usually with the aim of identifying information-theoretic limitations on the detection of  communities using   eigenvectors   \cite{nadakuditi2012hard,nadakuditi2013spectra,taylor2017super,taylor2016enhanced}. In this paper, we will bridge this theory to the application of  consensus learning.

We summarize  an approach developed in \cite{peixoto2013eigenvalue} for a block-structured random matrix $\tilde{{\bf X}}$ that is size $n$ and has $K$ blocks so that the expectation and variance of entries in a given block are the same, $\mathbb{E}[\tilde{X}_{ij}]=E_{rs}$ and $\mathbb{VAR}[\tilde{X}_{ij}]=V_{rs}$ for block $(r,s)$, and these can differ from block to block.
If we let ${\bf S}\in\mathbb{R}^{n\times K}$ be the one-hot encoding for which rows/columns belong to each block, then   the expectation and variance of ${\bf X}$ is given by
\begin{align}
    \mathbb{E}[\tilde{{\bf X}}] &= {\bf S}{\bf E} {\bf S}^T \\ \nonumber
    \mathbb{VAR}[\tilde{{\bf X}}] &= {\bf S} {\bf V} {\bf S}^T.
    \label{eq:var}
\end{align}
Note that both the expectation and variance are rank-$K$ matrices since ${\bf E}$ and ${\bf V}$ are size $K$.  

To proceed, it is convenient to define $\tilde{{\bf X}} ={\bf X}  + {\bf R} $ to separate $\tilde{{\bf X}}$ into its expectation ${\bf X} = \mathbb{E}[\tilde{{\bf X}}]$ and residual ${\bf R} =  \tilde{{\bf X}}-{\bf X}$. 
Because any eigenvalue $z$ of $\tilde{{\bf X}}$ solves $\det(z{\bf I} - \tilde{{\bf X}})=0$, one can use the property 
$z{\bf I} - \tilde{{\bf X}} = (z{\bf I} - {\bf R})({\bf I} - (z{\bf I} - {\bf R})^{-1}{\bf X}) $ to show that SBMs give rise two to types of eigenvalues:
\begin{itemize}
    \item \emph{bulk eigenvalues} that solve
    \begin{align} 
        \det \big(z\textbf{I} -\textbf{R}  \big) = 0; %\label{eq:bulky}
    \end{align}
    \item \emph{isolated eigenvalues} that solve
    \begin{align} 
        \det \big(\textbf{I}-(z \textbf{I} -  \textbf{R}  )^{-1} \textbf{X}  \big) = 0 .
        %\label{eq:isolated}
    \end{align} 
\end{itemize}
These two types are illustrated in Fig.~\ref{fig:RMT} for a normalized Laplacian matrix $\widehat{{\bf L}}$.  Note that $(  z \textbf{I} -  \textbf{R}  )$ is assumed to be invertible, which is guaranteed when $z$ is outside the spectral support of ${\bf R}$ (which is the defining feature of being `isolated').

Because the bulk eigenvalues of ${\bf X}$ are identical to those of ${\bf R}$,   they can be studied through the \emph{resolvent matrix}  $(z{\bf I} - {\bf R})^{-1}$ and the   Stieltjes transform \cite{benaych2011eigenvalues} 
$\rho(z) = -\frac{1}{n \pi} \text{Im}[ \text{Tr}(\mathbb{E}[(z{\bf I}-{\bf R})^{-1}])]$. It is argued in \cite{peixoto2013eigenvalue} that because the entries of ${\bf R}$ have blockwise-defined variances, which are identical to those for $\tilde{{\bf X}}$ as given by Eq.~\eqref{eq:var}, the resolvent matrix also has blockwise-defined properties enabling the following expression for the bulk spectral density
\begin{align} \label{eq:Stieltjes}
    \rho(z) 
    &= -\frac{1}{n \pi} \sum_{r=1}^K n_r \text{Im}[{t}_r(z)],
\end{align}
where  ${t}_r(z) =  [\mathbb{E}[(z{\bf I}-{\bf R})^{-1}]]_{ii}$ for $i \in \mathcal{V}_r$, which can be solved via a system of $K$ nonlinear equations for $r\in\{1,\dots,K\}$ 
\begin{align}  \label{eq:tr}
    t_{r}(z)  =  \sum_c \frac{P[R_{ii}=c]}{ z  - c - \sum_{s=1}^{K}  n_s t_{s}(z) V_{rs}  }.
\end{align} 
Here, $P[R_{ii}=c]$ denotes the probability that a diagonal entry in $R$ equals $c$.
If we let $f_r(z,{\bf t}(z))$ denote the right-hand side of Eq.~\eqref{eq:tr} and define ${\bf t}(z)=[t_1(z),\dots,t_K(z)]^T$ and $F(z,{\bf t}(z)) = [f_1(z,{\bf t}(z)),\dots,f_K(z,{\bf t}(z))]^T$, then for any $z\in\mathbb{R}$ the $t_{r}(z) $ values are an equilibrium of the nonlinear  fixed point iteration 
\begin{align}  \label{eq:it}
    {\bf t}^{(l+1)}(z) = F(z,{\bf t}^{(l)}(z))
\end{align} 
with $l=1,2,\dots$ and for some sufficiently accurate initial condition ${\bf t}^{(0)}(z)$.

Function $F$   can also be used to identify the support for the bulk distribution $\text{supp}(\rho)=\{z\in \mathbb{R}\,|\,\rho(z)\ge 0\}$, and in particular, the boundaries for $\text{supp}(\rho)$. Specifically, any solution ${\bf t}(z)$ to Eq.~\eqref{eq:it} will become unstable at such a boundary, in which   $\rho(z)=0$ and the Jacobian ${\bf J}(z)$ with entries $J_{rs}(z) = \frac{\partial f_r}{\partial t_s}$   satisfies 
 \begin{align} \label{eq:Jacobian}
    det \big( \textbf{I}_{r} - \textbf{J}(z) \big) = 0 .
 \end{align}
When the support of $\rho(z)$ is a single region, i.e.,
$\text{supp}(\rho)  = [\lambda_L,\lambda_R]$, then we use $\lambda_L$ and $\lambda_R$ to denote the left and right bounds on $\text{supp}(\rho)$.

Finally, we describe a technique in \cite{peixoto2013eigenvalue} to predict the isolated eigenvalues  for SBMs. Given a solution ${\bf t}(z)$ to Eq.~\eqref{eq:it}, we define ${\bf T}(z)=\text{diag}[{\bf t}(z)]$ to obtain  $\mathbb{E}[(z{\bf I} - {\bf R})^{-1}] = {\bf S}{\bf T}(z){\bf S}^T$.
We can the study Eq.~(2.8) in expectation to obtain  
\begin{align} \label{eq:iso_2}
    det \big( \textbf{I} -  \textbf{T}(z) {\bf E}{\bf N}  \big) = 0 ,
\end{align} 
where   ${\bf N} = \text{diag}[n_1,\dots,n_K]$ encodes the block sizes.
The left-hand side of Eq.~\eqref{eq:iso_2} is a polynomial in $z$ and its roots indicate the isolated eigenvalues, in expectation, which can be numerically obtained using a root-finding algorithm. Because each matrix in Eq.~\eqref{eq:iso_2}  is at most rank $K$, there are at most $K$ isolated eigenvalues, and it can be shown that there is a one-to-one correspondence between the eigenvalues of ${\bf E}$ and the expected eigenvalues of ${\bf X}$.

In Fig.~\ref{fig:RMT}, we illustrate  the distribution $\rho(\lambda)$ of bulk eigenvalues (left)  and  the isolated eigenvalues (right) for an unnormalized Laplacian matrix for an SBM with $K=4$ communities. Note that the red-colored curve and dots, respectively, accurately predict empirically observed values for the bulk and isolated eigenvalues. In this case, it is straightforward to show that for a normalized Laplacian, one has 
${\bf X}=\widehat{{\bf L}} = {\bf I} - {\bf D}^{-1/2}{\bf A}{\bf D}^{-1/2}$,
${\bf E} = \hat{{\bf D}}^{-1/2} {\bf \Pi}{\bf D}^{-1/2}$, and
${\bf V} = \hat{{\bf D}}  [{\bf \Pi}\circ(1-{\bf \Pi})\hat{{\bf D}} $, where $\hat{{\bf D}} = \text{diag}[\sum_s n_s\Pi_{rs}]$ encodes the expected node degree for nodes in each block $r\in\{1,\dots,K\}$. In addition, the diagonal entries in $\widehat{{\bf L}}$ are all one, so $P[R_{ii}=c]=\delta_1(c)$ in Eq.~\eqref{eq:tr} (i.e., the Dirac delta function).

\section{Impact of Community Structure on  Consensus Algorithm for Scalar Data}\label{sec:consensus}

Here, we present new theoretical and numerical insights for  how the convergence time $\tau_\epsilon$ for consensus dynamics is affected by community structure in random networks. We describe the experimental setup in Sec.~\ref{sec:c1}, our main findings in Sec.~\ref{sec:c2}, and we compare/contrast the scenarios of sparse and dense networks in Sec.~\ref{sec:c3}

\subsection{Experiment Design.}\label{sec:c1}
Using  stochastic block models (see Sec.~\ref{sec:SBM}), we constructed networks with varying prevalence of community structure. Specifically, we considered networks sampled from different SBMs in which we fixed the network size $n$, the sizes $n_r$ of the communities, and the edge probability $p_{in}$ for two nodes in the same community. We then constructed many networks while varing the probability $p_{out}\in[0,p_{in}]$ for edges between  nodes in different communities.

For each SBM, we computed random-matrix-theory-based predictions for the second-smallest eigenvalue $\lambda_2$ of the normalized Laplacian matrix $\widehat{{\bf L}}$ as well the  left bound $\lambda_L$ on the support for the bulk eigenvalues as described in Sec.~\ref{sec:RMT}. (Recall that  the convergence time satisfied $\tau_\epsilon\sim\mathcal{O}(1/\lambda_2)$ for the consensus dynamics described in Sec.~\ref{sec:vanilla}.)

Letting $\Delta = p_{in}-p_{out} \in [0,p_{in}]$ quantify the prevalence of community structure (i.e., $\Delta = 0$ corresponds to no communities, whereas $\Delta=p_{in}$ corresponds to when the communities are so strong that they are completely disconnected), we studied how $\tau_\epsilon$ and $\lambda_2$ depend on $\Delta$. The user-defined $\varepsilon$ parameter to determine convergence was set to be $10^{-10}$.

 \begin{figure}[t!]
  \centering
    \includegraphics[width=0.8\linewidth]{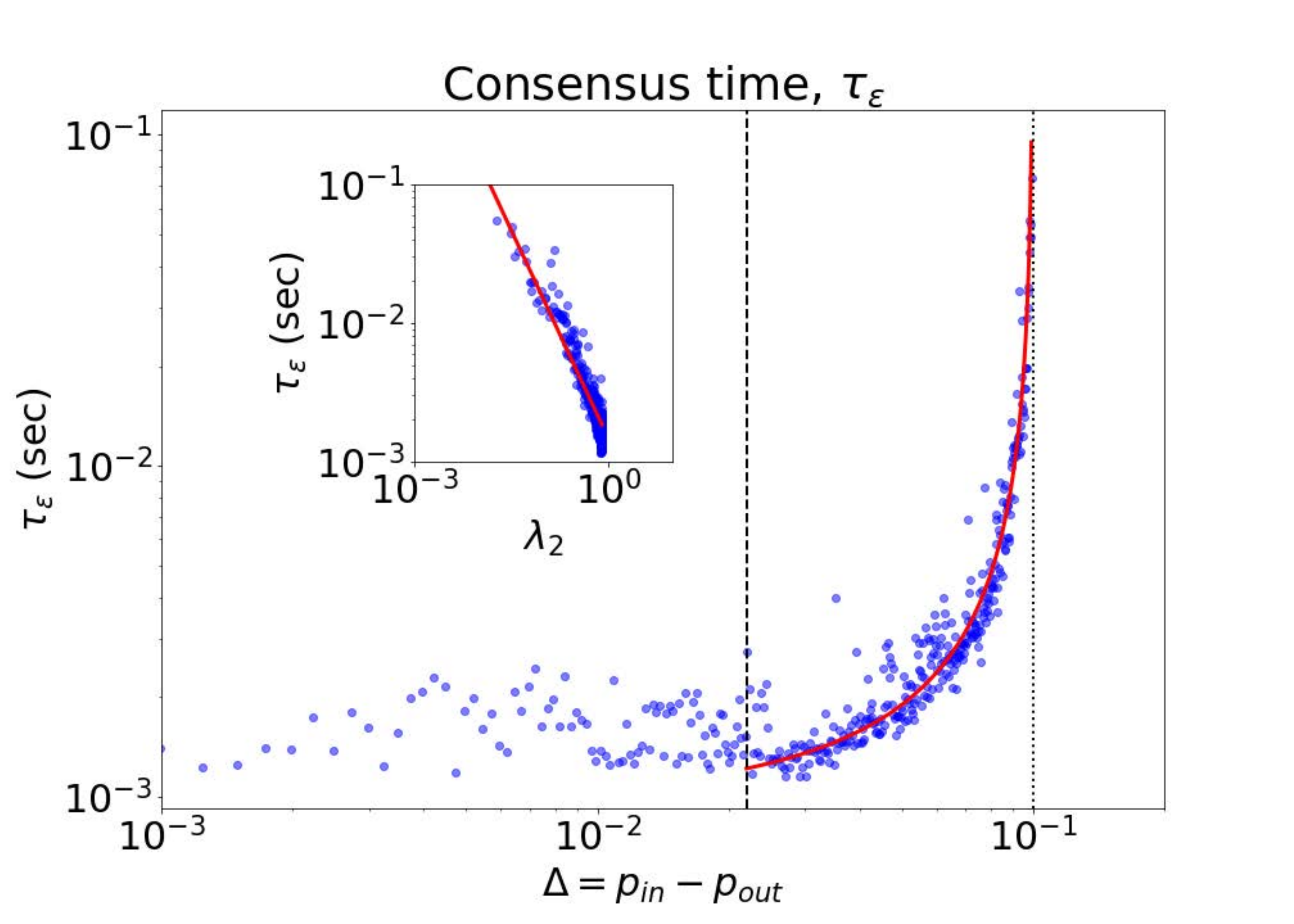}
    \includegraphics[width=0.8\linewidth]{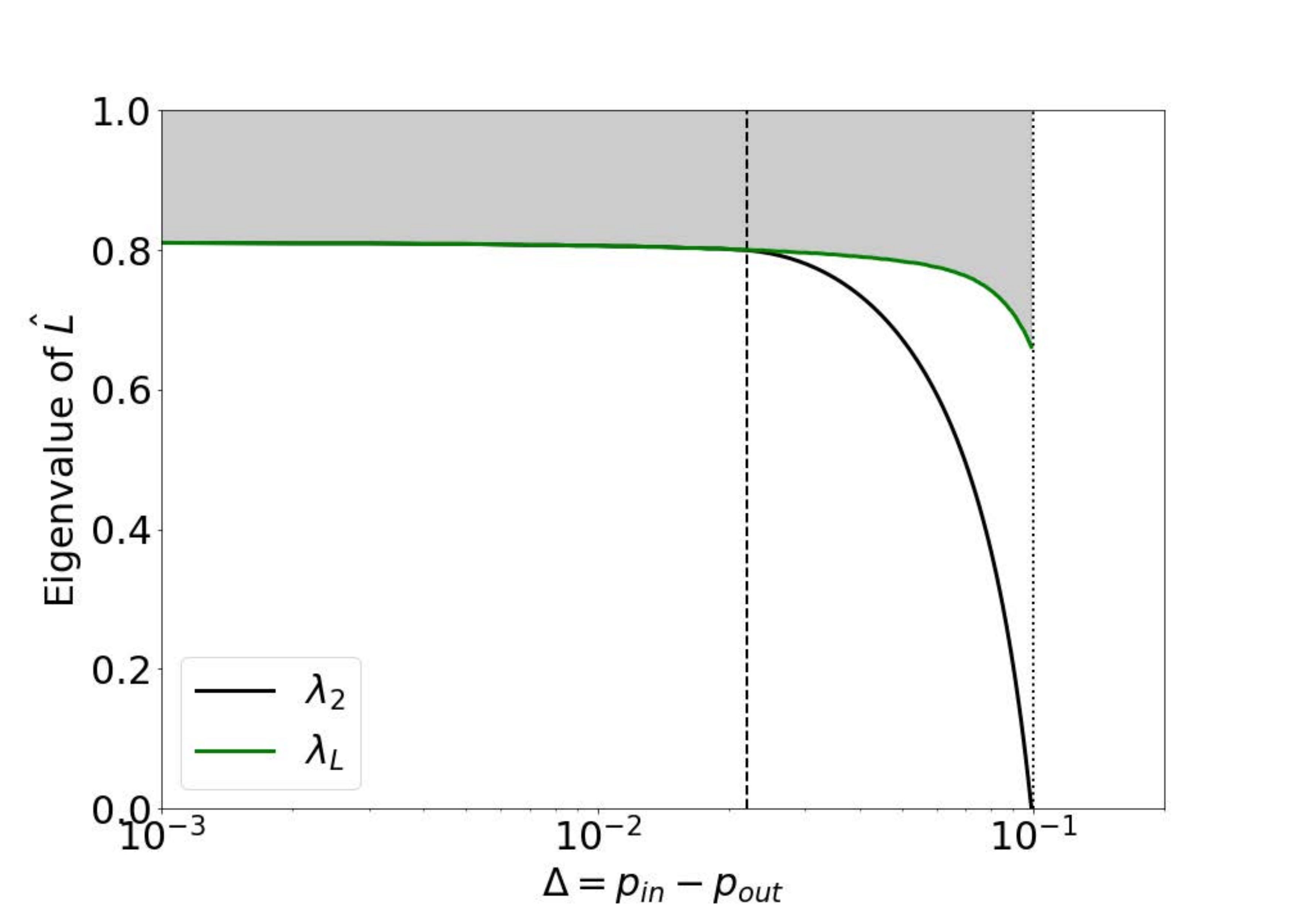}
    \caption{Consensus of scalar data over a sparser SBM  in which  $p_{in}=0.1$, and $p_{out}\in[10^{-3},0.1]$.
    (top panel) Empirical observations of convergence time $\tau_\epsilon$ (blue dots) versus $\Delta=p_{in}-p_{out}$.
    The red curve  is a line of best fit: $\tau_\epsilon \approx 0.000087/(0.1 - \Delta)$ The inset shows $\tau_\epsilon$ versus $\lambda_2$, and the red line indicates the fit $\tau_\epsilon\approx 0.0015/\lambda_{2}$.
    (bottom panel) 
    Predicted values for the second-smallest eigenvalue $\lambda_{2}$  of $\widehat{{\bf L}}$  and the left boundary $\lambda_{L}$ for the bulk spectral distribution $\rho(\lambda)$. 
    The gray shaded region indicates $supp(\rho)$. In both panels, vertical lines  indicate two critical values:
    a \emph{spectral bifurcation} occurs at $\Delta_1^*$ (dashed lines) in that the gap between $\lambda_2$ and $supp(\rho)$ disappears; $\tau_\epsilon$ diverges at $\Delta_2^*=p_{in}$ (dotted lines)  since $\lambda_2\to0$.
    }\label{fig:sparse}
\end{figure}

\begin{figure}[t!]
  \centering
    \includegraphics[width=0.8\linewidth]{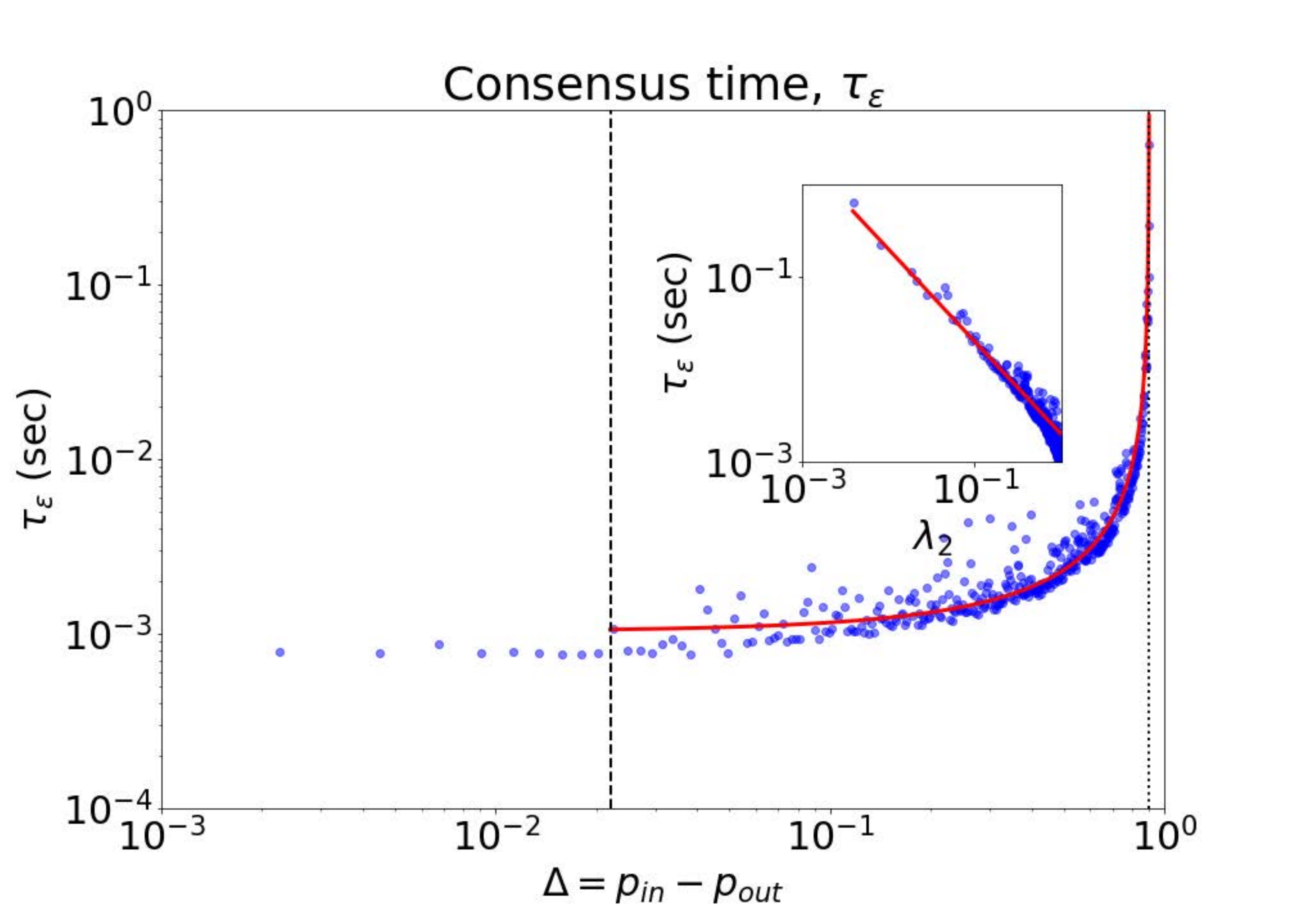}
    \includegraphics[width=0.8\linewidth]{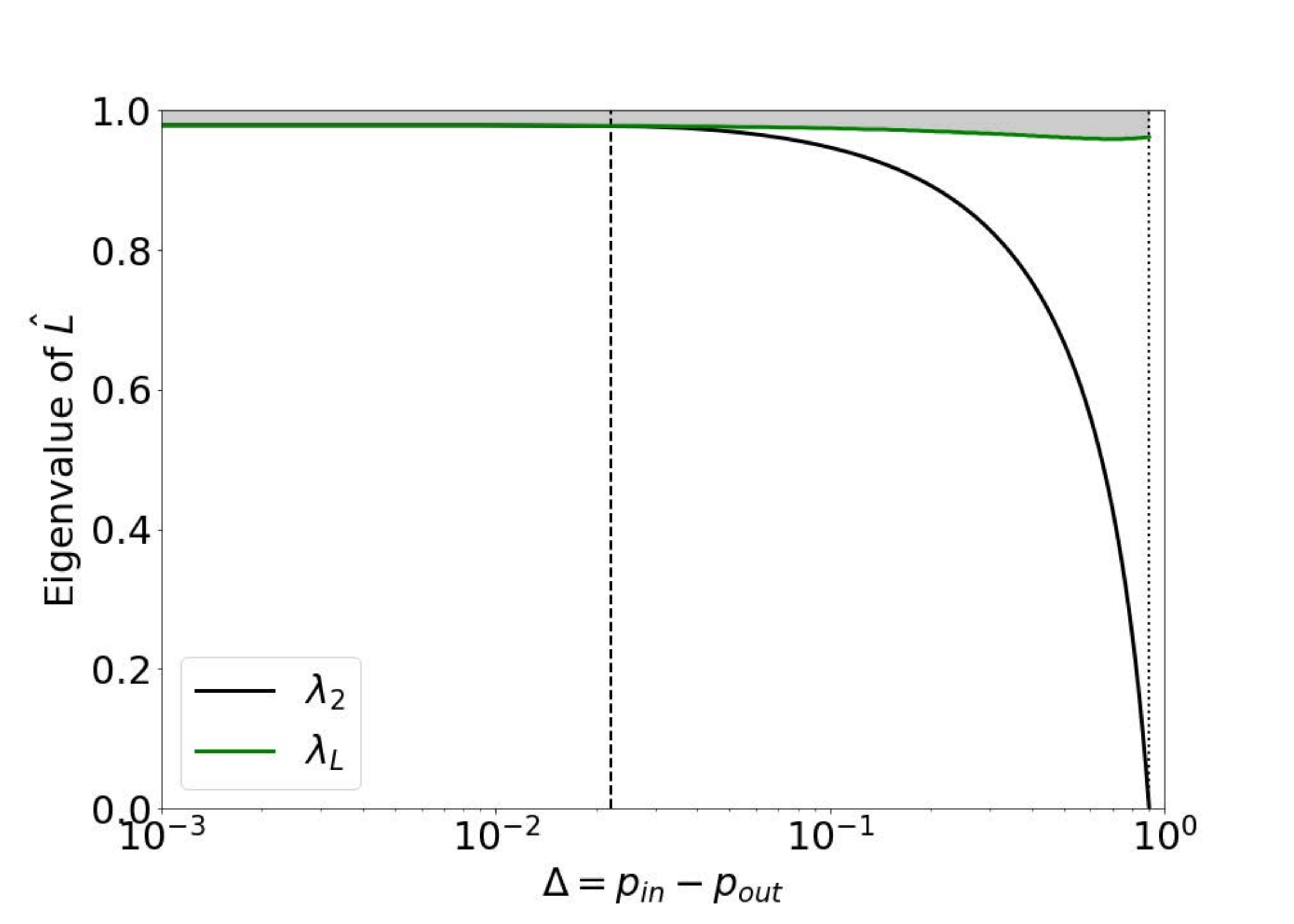}
   \caption{
    Consensus of scalar data over a denser SBM in which   $p_{in}=0.9$, and $p_{out}\in[10^{-3},0.9]$. The top and bottom panels are identical to those in Fig.~3, except the values are changed since the network now contains many more edges.
    (top panel) Note  when  $\Delta$ is very small (large) that the $\tau_\epsilon$ values are smaller (larger) that those for the sparser network as shown in Fig.~3(top). The lines of fit (i.e., red curves) are  $\tau_\epsilon\approx 0.0093/(0.9 - \Delta)$  and $\tau_\epsilon\approx 0.002/\lambda_{2}$.
    (bottom panel) Observe that $\lambda_L$ is now closer to 1 for all $\Delta$, indicating that $supp(\rho)$ is more concentrated near 1 for the denser network, which allows $1/\lambda_2$ to further decrease as $\Delta$ decrease.
    }
  \label{fig:dense}
\end{figure}

\subsection{Decreasing the Prevalence of Communities Speeds Consensus, to a Point.}\label{sec:c2}
In Fig.~\ref{fig:sparse}, we plot   empirically observed values of $\tau_\epsilon$ (top panel) and the analytically predicted second-smallest eigenvalue $\lambda_2$ of $\widehat{{\bf L}}$ (bottom panel). Results are shown for networks drawn from an SBM with $n=1000$ nodes, $K=2$ communities of sizes $[n_1,n_2]=  [700 ,300]$, $p_{in}=0.1$, and $p_{out}\in[10^{-3},0.1]$.
We refer to the networks as sparse since $p_{in}=0.1$, and we will later study denser SBMs with  $p_{in} = 0.9$.

Our first observation is that decreasing the extent of community structure (i.e., decreasing $\Delta$), causes $\lambda_2$ to increase and $\tau_\epsilon =\mathcal{O}(1/\lambda_2) $ to decrease. Hence, community structure generally inhibits consensus, and removing community structure  can decrease  $\tau_\epsilon$ by orders of magnitude. 

The vertical lines in Fig.~\ref{fig:sparse} indicate two critical values of $\Delta$, which provide a   more detailed understanding of this phenomenon.   The vertical dashed lines indicate $\Delta_1^*$, which is where a \emph{spectral bifurcation} occurs in that the gap disappears between $\lambda_2$ and $\lambda_L$, the left boundary of the bulk spectral density $\rho(\lambda)$ (whose support is indicated by the shaded region in the lower panel). Importantly, $\lambda_2$ and $\tau_\epsilon$ significantly vary with $\Delta$ when  $\Delta>\Delta_1^*$; in contrast, they are largely insensitive to $\Delta$ when   $\Delta<\Delta_1^*$.  The vertical dotted lines indicate $\Delta_2^*=p_{in}$, which is the value of $\Delta$ where $\lambda_2\to0$ and $\tau_\epsilon\sim \mathcal{O}(1/\lambda_2)$ diverges. (This occurs since $p_{out}\to0$ in this limit, and so the communities become disconnected components.)

\subsection{Denser Networks Are More Sensitive to Community Structure.}\label{sec:c3}
In Fig.~\ref{fig:dense}, we depict identical results as shown in Fig.~\ref{fig:sparse} except that we change $p_{in}$ from $0.1$ to $0.9$ so that the resulting networks contain many more edges. As a result, we can observe that the range of $\Delta \in [\Delta_1^*,\Delta_2^*]$ in which commmunity structure significantly affects $\tau_\epsilon$ is much wider. As a result, the  range of $\tau_\epsilon$ values is wider for the denser network than for the sparser network. In particular, for small $\Delta$ the $\tau_\epsilon$ values are smaller for the denser network than for the sparser network. This is largely due to the fact that $\rho(\lambda)$ concentrates near 1 more strongly for the dense network than for the sparse network (compare shaded regions in the lower panels of Figs.~3 and 4), allowing $\lambda_2$ to further increase as $\Delta $ decreases.

\section{Application to Distributed SVM Algorithm}\label{sec:gadget}
In recent years, a new interest has evolved around developing algorithms for on-board sensor networks (also called learning on edge devices). The Support Vector Machine (SVM) algorithm \cite{cortes_95a,Vapnik00} is one such algorithm that is used extensively in many commercial applications involving IoT devices and scalable variants of SVMs are well studied in literature \cite{Joachims_99a, Sindhwani_06a}.

In this section, we study how   community structures in sensor networks affects the convergence of a consensus SVM algorithm: the Gossip bAseD sub-GradiEnT Solver (GAGDET) SVM \cite{Hensel_09c,Dutta_18a,Dutta_20a}.   Below, we describe the algorithm and implementation (Sec.~\ref{sec:alg}), datasets
(Sec.~\ref{sec:data}), experiment design (Sec.~\ref{sec:exp}), and results
(Sec.~\ref{sec:res}).

\subsection{Algorithm and Implementation.}\label{sec:alg}
~

\noindent{\bf GADGET SVM.}
The goal of GADGET SVM is to learn a linear SVM on a global data set by learning local models at a set of nodes, and by exchanging information between nodes using a gossip-based consensus protocol.  In a decentralized setting, let $\mathcal{X}=\{{\bf x}^{(i)}\}\in\mathbb{R}^{d}$ denote a set of $n$ examples having $d$ features, and let $\mathcal{Y}=\{y^{(i)}\}$ be their class labels. Further, we \emph{horizontally  partition} the data across $n$  compute nodes/sensors by separating $\mathcal{X} $ and $\mathcal{Y}$  into $n$ disjoint subsets which we label $\mathcal{X}_i$ and $\mathcal{Y}_i$, respectively. (Note that $\mathcal{X}=\cup_i \mathcal{X}_i$ and $\mathcal{Y}=\cup_i \mathcal{Y}_i$.) We  study a primal formulation of the SVM problem using an objective function  ${J}( \textbf{w} ) $ defined by
\begin{align}  
    \arg \min_{ \textbf{w}  }   {J}( \textbf{w} ) = \frac{1}{n} \sum_{j=1}^{n} \sum_{{\bf x},y \in \mathcal{X}_j,\mathcal{Y}_j}  l \big( \textbf{w}; ( \textbf{x} ,y) \big) + \frac{\nu}{2} || \textbf{w} ||_{2}^{2}\label{eq:gadget}
\end{align} 
where $ l \big( \textbf{w}; ( \textbf{x},y) \big)$ is a loss function and $\frac{\nu}{2} || \textbf{w} ||_{2}^{2}$ is a regularization term with regularization constant $\nu$.

GADGET SVM is a decentralized algorithm to solve Eq.~\eqref{eq:gadget} that implements the Pegasos algorithm \cite{Shwartz_2007} at each node to obtain a node-specific weight vector ${\bf w}_j$.  At the beginning, ${\bf w}_j$ is set to the zero vector.  At iteration $t$ of the algorithm, a training example $(x_t^{(i)},y_t^{(i)})$ is chosen uniformly at random and the sub-gradient of the objective is estimated. At iteration $t+1$, the local weight vector is updated using a Stochastic Gradient Descent (SGD) step as follows: ${\bf w}_j^{t+1} = {\bf w}_j^{t} - \eta_j^t \triangledown_j^{t}$, where $\triangledown_j^{t}$ is the sub-gradient of the objective function and $\eta_j^t$ is the learning rate at iteration $t$.

At the same time, a gossip based communication protocol called Push Sum \cite{Kempe_03a} is used to exchange the learned weights ${\bf w}_j$ between nodes through a communication network. We use a modified version in which nodes share/store weight vectors, and we briefly summarize a scalar version of the algorithm below. In summary,  each node $i$ stores a sum $s_{t,i}$ and weight $w_{t,i}$, with the initial weights $w_{0,i}=1$. For $t=0,1,2,\dots$,  some neighbors $j_{1}, \hdots, j_{k} \in \mathcal{N}(i)$ send the set of pairs $\{ \big(  s_{ j_{l},t } , c_{ j_{l},t  } \big) \}_{l=1}^{k}$ to node $i$, which updates its sum and weight via $s_{t+1,i} = \sum_{l=1}^{k} \textbf{P}_{ j_{l} i } s_{  j_{l}, t }$ and $w_{t+1,i} = \sum_{l=1}^{k} \textbf{P}_{ j_{l} i } w_{  j_{l}, t }$, where $\textbf{P}\in\mathbb{R}^{n\times n}$ is  a row-stochastic transition matrix that encodes the communication networks. Each entry $\textbf{P}_{ij}$ specifies the probability of a communication between node $i$ and $j$. Finally, after updating, node $i$ sends its new values to one of its senders, and the process continues.

\noindent\textbf{Implementation.} 
GADGET SVM has been implemented on Peersim \footnote{\url{http://peersim.sourceforge.net/}}, a peer-to-peer network(P2P) simulator. This software allows simulation of the network by initializing sites and the communication protocols to be used by them. GADGET implements a cycle driven protocol that has periodic activity in approximately regular time intervals. Sites are able to communicate with others using the Push Sum protocol. The experiments are performed on a laptop equipped with 1.6 GHz Intel core i5 dual core processor; main memory size of 256 GB and 4GB RAM.

\subsection{Datasets.}\label{sec:data}
We report results for the following benchmark datasets.
\begin{itemize}
    \item Quantum  (obtained from  \url{http: //osmot.cs.cornell.edu/kddcup/}) 
    The dataset has 50000 (40000 training)  labeled examples with 78 attributes  generated in high energy collider experiments. There are two classes representing two types of particles.
    
    \item MNIST (obtained from   \url{http://yann.lecun.com/exdb/mnist/}) 
    The dataset has 70000 (60000 training) labeled images with 123 features, which represent a compressed form of images that are size $28 \times 28$. The binary classification task is to predict whether  or not a character is an 8.

\end{itemize}

\subsection{Experiment Design.}\label{sec:exp}
The following steps were performed on each dataset.

First we sampled  an SBM to obtain a network containing $n$ nodes and $K=2$ communities of sizes $n_1=30$ and $n_2=70$, with within-community and between-community edge probabilities given by $p_{in}=0.9$ and various $p_{out}$ We considered 15 different choices for $p_{out} \in [10^{-3},p_{in}]$.  

Then, we imputed each network into the Peersim simulator to implement the GADGET algorithm over it. At the same time, we split the training and testing sets into $n$ different files, containing approximately equal numbers of examples, which we then distributed amongst the nodes. Next, the GADGET algorithm was executed on each node independently until the local weight vectors were $\varepsilon-$close to each other, i.e., $|| \textbf{w}_{i} - \textbf{w}_{j} ||_{2}<\varepsilon$ for all node pairs $i,j$. We set $\epsilon=10^{-10}$. The local models are then used to determine the primal objective and the test error on the corresponding test sets. This entire process was executed several times, and   average values of the primal objective and test error were obtained.

\subsection{Results.}\label{sec:res}
\begin{figure}[t!]
  \centering
    \includegraphics[width=0.9\linewidth]{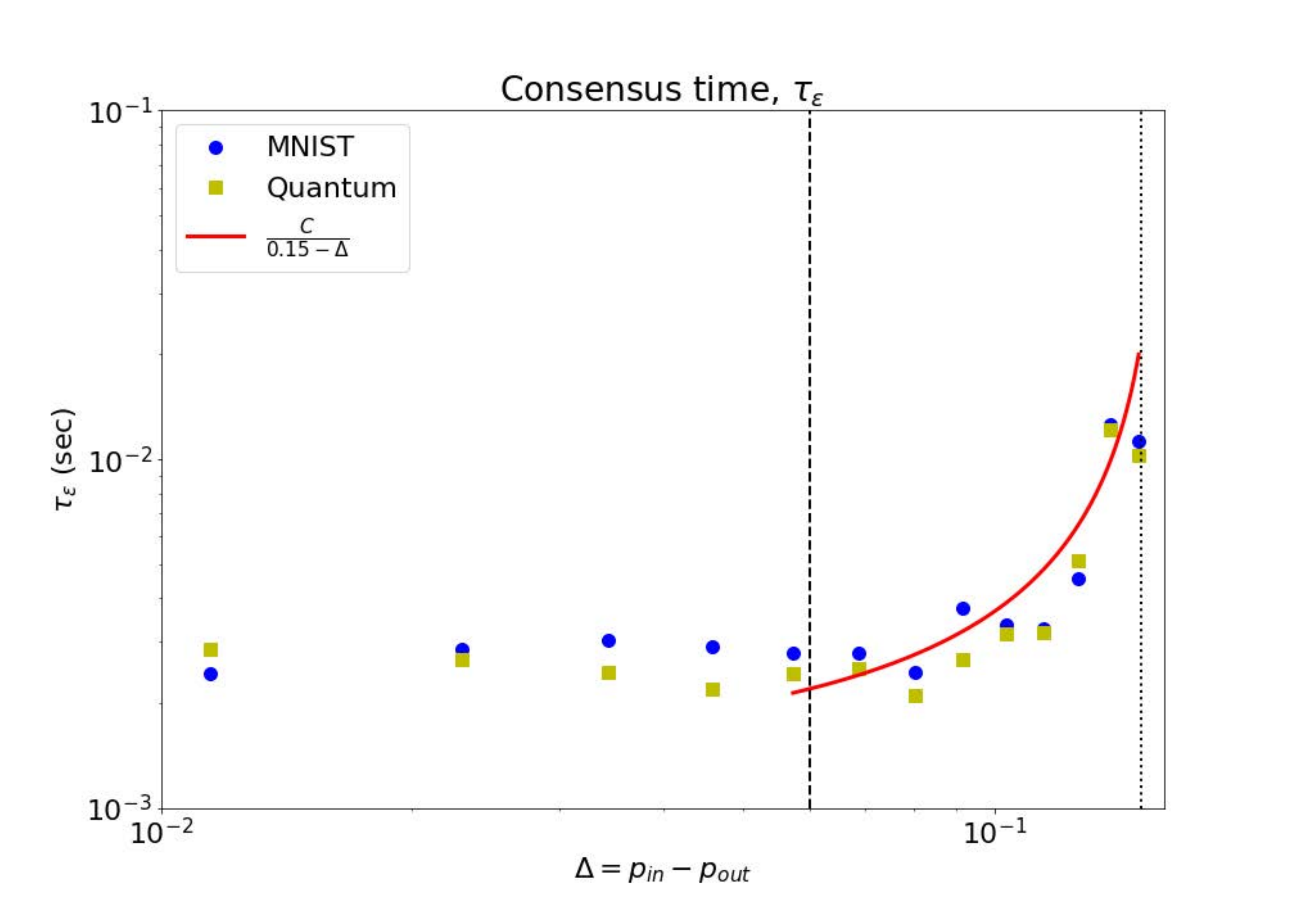}
    \includegraphics[width=0.9\linewidth]{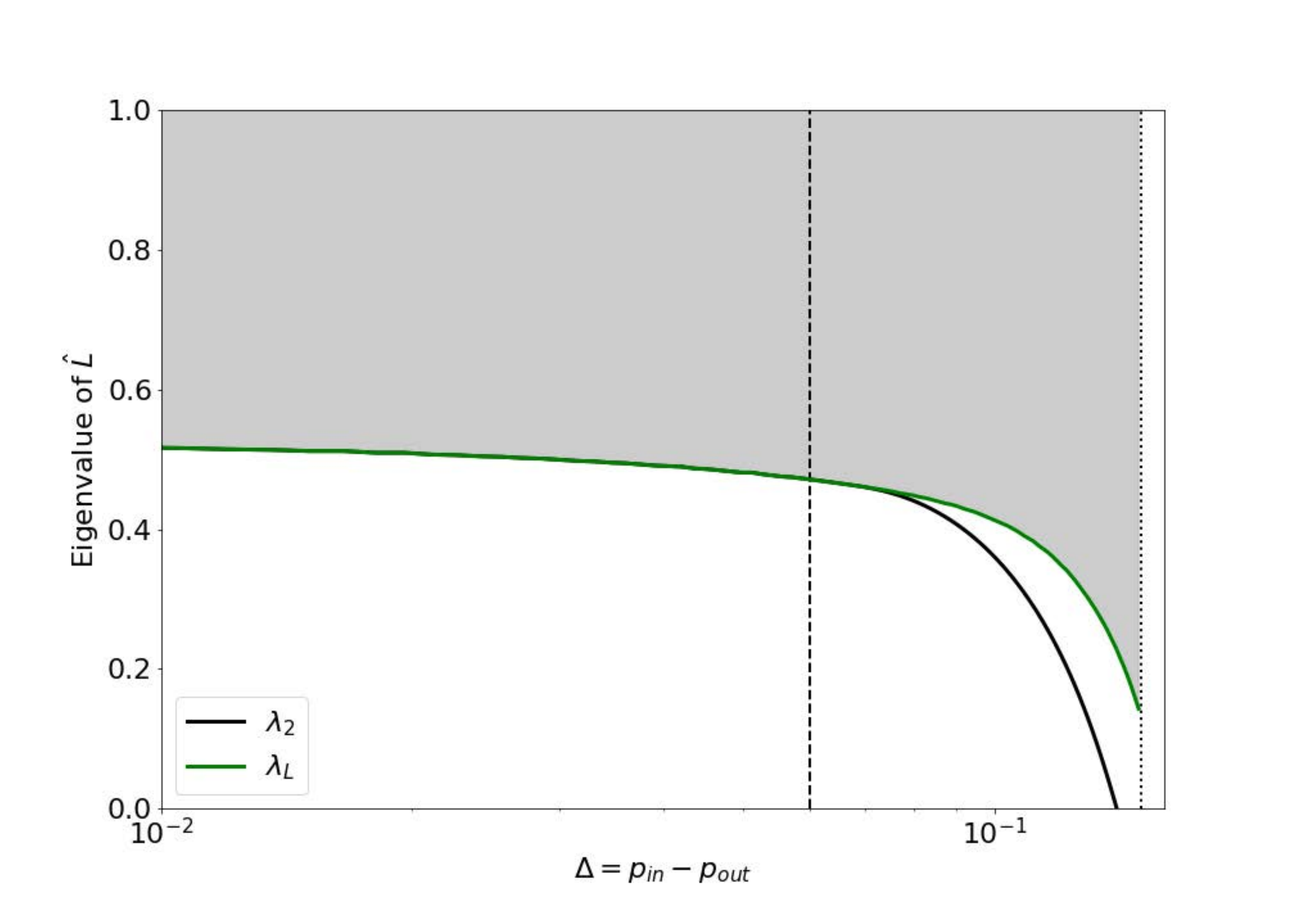}
    \caption{Impact of community structure on decentralized SVM algorithm GADGET.
    (top panel) Consensus time versus community strength $\Delta$ for the application of Gadget to two datasets: MNIST  and Quantum. 
    (bottom panel) Eigenvalues $\lambda_{2} $ and $ \lambda_{L}$, as obtained from random matrix theory, as shown versus $\Delta$. Similar to our findings in Sec.~3, $\tau_\epsilon$ is significantly dependent on community structure when $\Delta\in[\Delta_1^*,\Delta_2^*]$ (as indicated by the dashed and dotted vertical lines). The red curve is a line of best fit $\tau_\epsilon\approx {0.00022}/{0.15-\Delta}$. 
    }\label{fig:sparse5}
\end{figure}

In Fig.~\ref{fig:sparse5}, we present figures that are similar to Figs.~3 and 4, except that we plot $\tau_\epsilon$ and the eigenvalues $\lambda_2$ and $\lambda_L$ versus $\Delta$ for the decentralized SVM algorithm. Our main finding is a qualitative similarity in the underlying property that $\tau_\epsilon$ strongly depends on community structure in an SBM when $\Delta$ lies within some range $[\Delta_1^*,\Delta_2^*]$, which were defined in Sec.~3.
Note that despite the two datasets being very different (MNIST  and Quantum), their convergence behavior appears to be qualitatively similar. It's worth noting that we found the classification accuracy of the GADGET algorithm after consensus largely did not vary with community structure, and that their values were comparable to other decentralized SVM implementations for these datasets ($\approx 72\%$ for MNIST and  $\approx 56\%$ for Quantum) \cite{Flouri_09a}.

\section{Conclusion}\label{sec:conclusion}

In this paper, we employed eigenvalue analysis techniques for SBMs to study the effect of community structure on consensus based learning. Analysis for consensus learning on randomly generated scalar data showed that consensus time $\tau_{\epsilon}$ has a reciprocal relationship with the community strength $\Delta=p_{in} - p_{out}$. Our numerical experiments with GADGET SVM confirmed this reciprocal relationship between $\tau_{\epsilon}$ and $\Delta$ for decentralized support vector machines. Furthermore, in both settings we observed two regimes in which $\tau_\epsilon $ varies significantly with $\Delta$ when $\Delta$ is larger (i.e., community structure is very strong), and is largely independent of $\Delta$ when $\Delta$ is sufficiently small. Our techniques and insights can aid the design and optimization of consensus-based machine learning algorithms over random networks that contain community structure as a result of avoiding long-range communications.

%In future work, ...
%According to David Kempe, the Push Sum procedure alone has consensus convergence time being $\mathcal{O} \Big( \frac{1}{  \lambda_{2}( \widehat{ \mathbfcal{L} } ) }  \Big)$. According to Bijral, Sarwate, and Srebro in \textcolor{red}{Data-Dependent Convergence for Consensus Stochastic Optimization}, the convergence process of a non Push Sum consensus-based SGD algorithm is also $\mathcal{O} \Big( \frac{1}{  \lambda_{2}( \widehat{ \mathbfcal{L} } ) }  \Big)$. Our numerical results above show that the combination of Push Sum and SGD is $\mathcal{O} \Big( \frac{1}{  \lambda_{2}( \widehat{ \mathbfcal{L} } ) }  \Big)$. The rigorous proof of this result will be shown in a future paper.

%%%%%%%%%%%%%%%%%%%%%%%%%%%%%%
% bibliography
%%%%%%%%%%%%%%%%%%%%%%%%%%%%%%
\bibliographystyle{siam}
\bibliography{RMT_refs}
\end{document}